\documentclass[twocolumn]{svjour3}          
\usepackage{graphicx}
\usepackage{natbib}

\usepackage{enumitem}
\usepackage{pifont}
\usepackage{algorithm, algorithmic}
\usepackage{amsmath}
\usepackage{caption}
\usepackage{subfig}

\usepackage{xcolor}
\usepackage{array}
\begin{document}
\graphicspath{{./Assets/}}
\title{Graph Based Over-Segmentation Methods for 3D Point Clouds
}


\author{Yizhak Ben-Shabat\and Tamar Avraham\and Michael Lindenbaum \and \\Anath Fischer}


\institute{ Y. Ben-Shabat \at
            Mechanical Engineering Department, Technion - Israel Institute of Technology, Haifa 32000, Israel. \\
            Tel.: +972-4-829-2334 \\
            \email{ sitzikbs@campus.technion.ac.il}
                       \and
           T. Avraham \at
            Computer Science Department, Technion - Israel Institute of Technology, Haifa 32000, Israel. \\
            Tel.: +972-4-829-4877 \\
            \email{ tammya@cs.technion.ac.il}
                       \and
           M. Lindenbaum \at
            Computer Science Department, Technion - Israel Institute of Technology, Haifa 32000, Israel. \\
            Tel.: +972-4-829-4331 \\
            Fax.: +972-4-829-3900 \\
            \email{ mic@cs.technion.ac.il}
           \and
           A. Fischer \at
            Mechanical Engineering Department, Technion - Israel Institute of Technology, Haifa 32000, Israel. \\
            Tel.: +972-4-829-3260 \\
            Fax.: +972-4-829-5711 \\
            \email{ meranath@technion.ac.il}
}

\date{Received: date / Accepted: date}

\maketitle

\begin{abstract}

Over-segmentation, or super-pixel generation, is a common preliminary stage for many computer vision applications. New acquisition technologies enable the capturing of 3D point clouds that contain color and geometrical information. This 3D information introduces a new conceptual change that can be utilized to improve the results of over-segmentation, which uses mainly color information, and to generate clusters of points we call super-points. We consider a variety of possible 3D extensions of the Local Variation (LV) graph based over-segmentation algorithms, and compare them thoroughly. We consider different alternatives for constructing the connectivity graph, for assigning the edge weights, and for defining the merge criterion, which must now account for the geometric information and not only color. Following this evaluation, we derive a new generic algorithm for over-segmentation of 3D point clouds. We call this new algorithm Point Cloud Local Variation (PCLV). The advantages of the new over-segmentation algorithm are demonstrated on both outdoor and cluttered indoor scenes. Performance analysis of the proposed approach compared to state-of-the-art 2D and 3D over-segmentation algorithms shows significant improvement according to the common performance measures.

\end{abstract}
\begin{sloppypar}
\section{Introduction}
\label{Sec:Intro}

    \subsection{Overview}
    \label{SubSec:Overview}

    Image segmentation methods aim to divide the input data according to object association. This inverse problem is often ill-posed. It is therefore common to relax the requirement, and aim for an over-segmentation in which the number of segments can be greater than the number of objects in the scene. The goal is then to get a rather small set of segments that do not cross object boundaries; an over-segmentation algorithm is considered to perform well when the segment boundaries overlap with the ground truth object boundaries while also partially covering only a single object's area. Over-segmentation can be considered as a compact and informative description of the scene, using a substantially lower number of elements than the initial pixel representation. It significantly reduces the amount of data that must be dealt with for the following image analysis stages, without loss of information.  The output elements of the over-segmentation process are commonly called super-pixels. Super-pixels were shown to be very useful as inputs for higher-level vision tasks such as  semantic segmentation \citep{Gould2008a, Farabet2013}, scene classification \citep{Juneja2013}, 3D geometry inference \citep{Hoiem2007}, and tracking \citep{Oron2014,ShuWang2011}.

    In recent years, 3D point cloud representation of geometric real-world data has gained popularity in the fields of robotics, computer vision, and computer aided design. This is mainly due to the emergence of low cost 3D sensing devices such as the Microsoft Kinect for indoor scenes and the more expensive LiDAR for outdoor scenes. Many point cloud based algorithms are being developed for applications such as navigation, object recognition, pose estimation, registration, and surface reconstruction. One of the main challenges when working with 3D point clouds is the large number of points. This challenge becomes even greater for real-time applications. One way to deal with the large amount of data is to divide it into subsets and process them separately. The data can be subdivided by a 3D regular grid, but data-driven subdivision that relies on 3D properties might be much more effective.

    Many over-segmentation methods have been proposed for 2D images, while very limited research has been done on adapting the over-segmentation approach for 3D point clouds. We focus here on extensions of one of the leading 2D over-segmentation algorithms, Local Variation (LV) by \cite{Felzenszwalb1998,Felzenszwalb2004}. A few works incorporated 3D extensions for it in different applications. However, none of these works compared the proposed extension to possible alternatives. Here we propose, for each step of the original algorithm, options for extending it to 3D point clouds, and discuss the pros and cons of each. We thoroughly test and compare each option and select the best combination. We call this new algorithm Point Cloud Local Variation (PCLV). Like the original algorithm, PCLV is simple and fast yet powerful. The algorithm is generic and can be applied on any data represented by a 3D point cloud, regardless of its geometrical complexity or the sensor used for its acquisition.  We show that PCLV performs better than state-of-the-art 2D methods and better than all previously suggested 3D methods on an extensive benchmark of indoor Kinect data. Additionally, we demonstrate its performance on urban scenes captured by an interferometry sensor.
    Our main contributions are:

    \begin{itemize}[label=\textbullet]
      \item  A class of 3D over-segmentation algorithms, obtained by applying the LV principles with different choices.
      \item  The PCLV algorithm, obtained by optimizing over all choices, which outperforms state-of-the-art 2D and 3D methods.
      \item  A thorough performance analysis of 3D point cloud extensions of the LV algorithm.
      \item  A performance analysis of existing 2D and 3D over-segmentation algorithms on a large indoor dataset.
      \item  Qualitative and promising results for new outdoor scenes.
    \end{itemize}

    The paper is structured as follows. The rest of this section overviews  existing 2D and 3D over-segmentation algorithms. Section \ref{Sec:3D_Extensions} presents the 3D point cloud over-segmentation algorithm analysis, and its results are summarized in Section \ref{Sec:Evalutation}. The PCLV algorithm is presented in Section \ref{Sec:PCLV}‎. Experimental results and comparison to state-of-the art algorithms are presented in Section \ref{Sec:PerformanceAnalysis} along with some implementation details. Finally, Section \ref{Sec:Discussion} discusses the results and concludes.

    \subsection{Related Work}
    \label{SubSec:Related}
    In this section, we briefly overview existing over-segmentation methods for 2D images and 3D point clouds.
    \subsubsection{2D Over-Segmentation Methods (Super-Pixel Generation)}
    \label{SubSec:2D_OS}
    A short survey and comparison of existing 2D over-segmentation methods can be found in \cite{Achanta2012}. Classic algorithms include  Graph-cut \citep{Malik2000},  Meanshift \citep{Comaniciu2002}, and  Watershed \citep{Meyer1994}. Current state-of-the-art over-segmentation methods include the Turbopixels algorithm (TP) \citep{Levinshtein2009a}, which evolves a set of curves that converge to the segment borders using a level set approach, and the Simple Linear Iterative Clustering (SLIC) algorithm \citep{Radhakrishna2010,Achanta2012}, which is based on the k-means clustering algorithm.
    One of the best performing yet efficient over-segmentation methods is the graph based Local Variation (LV) algorithm \citep{ Felzenszwalb1998, Felzenszwalb2004}. A connectivity graph is constructed by connecting each pixel to its 4 or 8 closest neighbors. The square root of the squared sum of differences in color space between connected pixels is assigned as the weight on the graph edges.  Then the algorithm iteratively merges pixel regions by comparing the dissimilarity between regions to the internal dissimilarity of pixels within the region. The decision rule used in the LV algorithm was recently explained, in a probabilistic setting, as hypothesis testing from which the Probabilistic Local Variation (pLV) algorithm \citep{Baltaxe2015} was derived. Other 2D variations of LV have been suggested \citep{Fahad2006,Zhang2006}, as well as video-based extensions \citep{Grundmann2010}.
    In this work we thoroughly analyze the 3D extension of each LV algorithm stage and propose a new over-segmentation algorithm for 3D point clouds.

    \subsubsection{3D Over-Segmentation Methods (Super-Point Generation)}
    \label{SubSec:3D_OS}
    Similarly to 2D over-segmentation algorithms,which divide an image into segments, 3D over-segmentation methods divide a 3D point cloud into clusters, which we call super-points. The desired properties of the super-points are the same as for super-pixels:  a super-point's boundaries should include the object boundaries and overlap with only one object, while the total number of super-points should be small.
    In contrast to over-segmentation of 2D images, little work has been done on over-segmentation of 3D point clouds.

    The Voxel Cloud Connectivity Segmentation (VCCS) algorithm \citep{Papon2013} followed the 2D SLIC \citep{Achanta2012} algorithm and uses a variant of k-means clustering for the labeling of points. First, the cluster seeding is done by partitioning the 3D space into a regular grid of voxels. Second, strict spatial connectivity is enforced by the iterative clustering algorithm. The similarity between candidate points is defined by a distance function that takes into account the spatial distance, color and FPFH descriptors \citep{Rusu2009d} between two points. This algorithm has the advantage of producing regular and uniform sized segments. However, it has five adjustable parameters: the initial voxel grid resolution, the seed voxel grid resolution, and three importance factors, one for each of the color, distance, and geometrical components. A main disadvantage of this algorithm is its dependence on these parameters, which create a tradeoff between accuracy and the running time of the algorithm.

    The Boundary-enhanced Supervoxel Segmentation (BESS) algorithm \citep{Song2014} starts with a pre-processing boundary detection stage and applies a clustering process on a connectivity graph that excludes these boundary points. The geometric features used in the clustering step are the spatial coordinates concatenated with the angles, and angle distribution between horizontal and vertical neighboring points. The main drawback of this algorithm is its reliance on the point order. It assumes that the input is point data from LiDAR with horizontal and vertical consecutive points. This constraint is used both in the adjacency graph construction and in the boundary detection stage, making this algorithm inapplicable to general, unorganized point clouds.

    3D LV extensions were proposed in three papers. Each extension used a different graph construction method, different modalities to quantify the dissimilarity between neighboring points, and a different merge criterion for the clustering of sub-graphs. It remains unclear how each of these differences affected overall algorithm performance. \cite{Strom2010} used a sensor dependent mesh for the connectivity graph, and weighted the edges by both color dissimilarity and normal direction differences. Their merge criterion considered the two dissimilarities separately. \cite{Schoenberg2010} combined these dissimilarities along with the Euclidean distance into a single weight. \cite{Karpathy2013}  omitted the color information and used an assumption that concave regions are related to boundaries; their weight included a penalty for the dissimilarity of angle between normal vectors in these regions. These three papers focus largely on the sensor and system setup. In two of them, over-segmentation is not an end in itself but a tool for achieving other goals. The evaluation process was either qualitative or in terms of the end goal. Our focus here is on over-segmentation as an end in itself. We present several 3D variations to this algorithm and evaluate the performance of these variants quantitatively and qualitatively.

\section{Local Variation (LV) 3D Extensions}
\label{Sec:3D_Extensions}
In this work we address the problem of achieving an accurate over-segmentation result in 3D point clouds using a graph based approach. The input of the method is a 3D point cloud representation of the boundary surface of any general object or scene. The output is a division of the data into small segments. These segments are essentially clusters of proximate points sharing geometric properties and appearance, which are different from points in surrounding clusters.
The original LV algorithm and all the 3D extensions discussed here are constructed of the following steps:
\begin{enumerate}
  \item Graph construction
  \item Descriptor computation
  \item Edge weight assignment
  \item Sequential  subgraph criteria based merging
\end{enumerate}

\begin{figure}
\centering
  \includegraphics[width=0.5\textwidth]{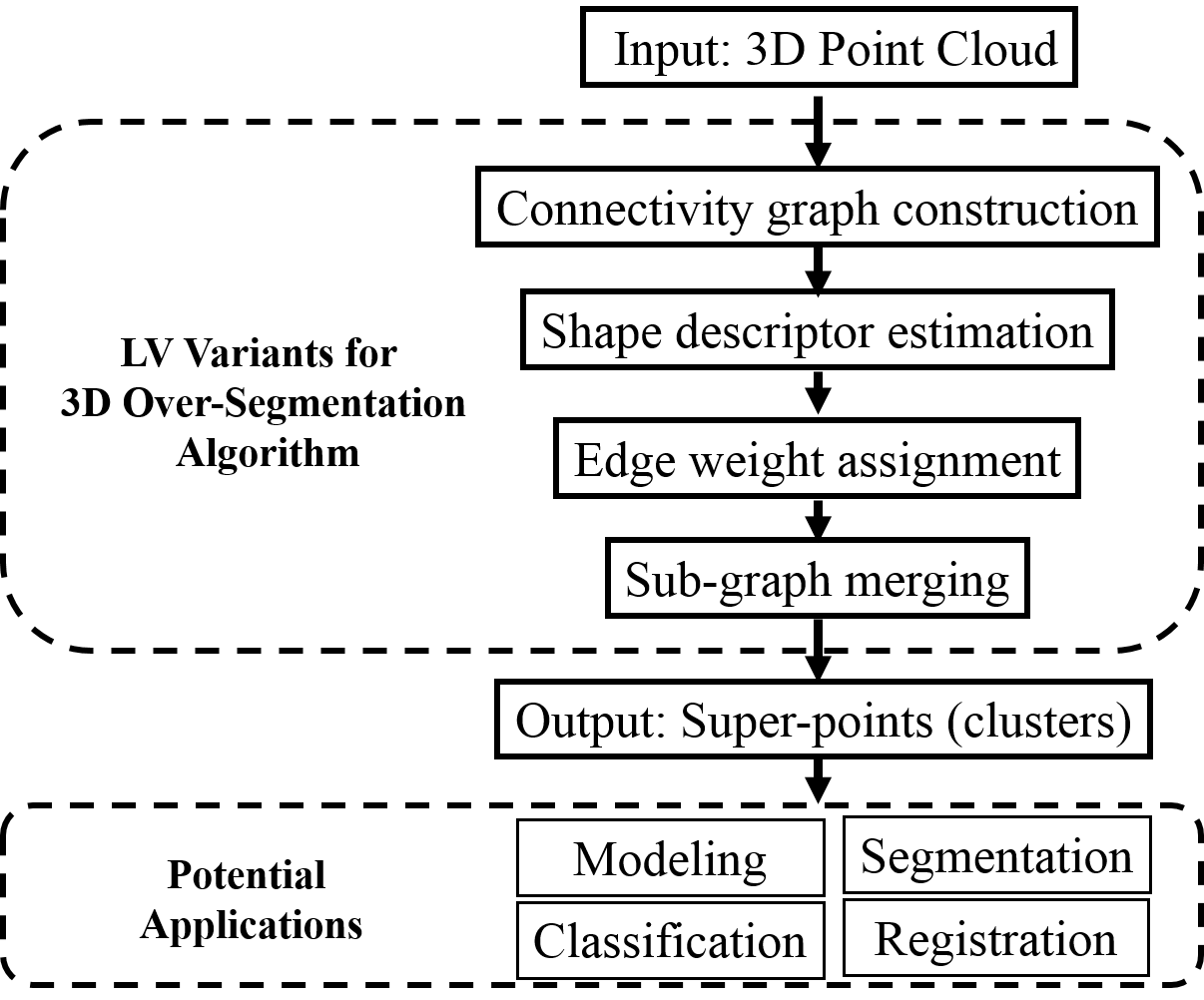}
\caption{Block diagram of LV variants for 3D over-segmentation}
\label{fig:BlockDiagram}       
\end{figure}

In the first step a graph is constructed from the input 3D point cloud data. In the second step, for each 3D point, a descriptor is estimated. The descriptor is a quantity, usually a vector of several elements, which characterize the point and its local environment. Some elements may be appearance related e.g. color, intensity, or local texture, and some may relate to local geometrical properties of the point with respect to its neighbors. The third step assigns one weight or more to each graph edge. The weight is a dissimilarity measure between two connected points. While the dissimilarities arising from different properties are often combined into one scalar weight, using an ``adding apples to oranges'' approach, we use a different approach to tackle the challenge of combining these properties, which we will refer to henceforth as modalities. The core of the over-segmentation method is the final merge stage.  All input points are initially super-points, which are iteratively merged to form larger super-points: the algorithm sorts the graph edges in ascending order and then traverses over the edges and applies a decision criterion to determine whether the super-points connected by the edge should be merged. Figure \ref{fig:BlockDiagram} summarizes the main stages of the approach.
Next, we detail the factors that must taken into account when choosing how to extend each of the stages for 3D point clouds.
    \subsection{Graph Construction}
    \label{SubSec:GraphCon}
    The construction of the connectivity graph $G = (V,E)$ is a crucial stage in the algorithm. It essentially defines the space of possible clusters. If there is no path between two points in the connectivity graph, they will never be in the same final cluster. However, too many graph edges decrease the algorithm's efficiency. We consider four alternative methods for constructing the connectivity graph for a 3D point cloud:
    \begin{enumerate}
      \item Connecting every two points whose distance from each other is less than a threshold R.
      \item Connecting each point to its K-nearest neighbors.
      \item Constructing a Delaunay triangulation and using it as the connectivity graph.
      \item Using a 2D image grid to construct a 4-connected or 8-connected graph, when such a projection is available (as originally constructed by the LV algorithm).
    \end{enumerate}
    There are advantages and drawbacks to these different options.

Constructing a graph using neighbors within a given radius R limits the length of graph edges, filtering out irrelevant connections. However, it also creates disconnected components for isolated points or points in sparse regions, actually performing a segmentation decision that was originally assigned to later stages in the algorithm. In addition, the number of points encapsulated in each sphere may vary drastically between different sections of the point cloud, creating unbalanced graphs and a bias toward dense regions. Most importantly, a majority of real-world point clouds have internally varying point densities, which makes it hard to adjust the R parameter.

The K-nearest neighbor graph overcomes some of these drawbacks: it is less sensitive to varying point cloud densities. In addition, the number of neighbors for each point is predetermined, eliminating the bias and creating a balanced graph. However, the K parameter still requires adjustment. This parameter has a significant impact on speed due to the required K-nearest neighbor search. Furthermore, distant non-related points may be connected in this graph.

The Delaunay mesh approach requires a triangulation stage that is sensitive to noise as outlier points may result in mesh structure that does not represent the underlying geometry. Furthermore, similarly to the K-nearest neighbor approach, triangulation may create synthetic connections between distant points.

Using a 2D grid based graph requires a mapping between the point cloud and an image. This mapping is easily obtained when working with low cost sensors such as the Microsoft Kinect, but is not always available for other sensors. Once the image is given, the connectivity graph is given as well, making the graph construction stage highly efficient in comparison to the other graph construction methods. Furthermore, although this method does not directly facilitate the geometry of all three dimensions, it still exploits a planar geometric relationship between the sampled points in a projection plane.

\begin{figure}
\centering
  \includegraphics[width=0.2\textwidth]{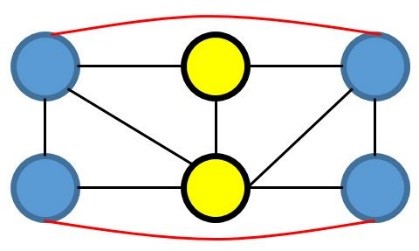}
\caption{Dispersed segments. Blue and yellow nodes: points in two different super-points; black lines: graph edges of an 8-connected image-based graph; red lines: additional edges used by other graph construction methods.}
\label{fig:DisperdedSegIllustration}
\end{figure}

\begin{figure}
\centering
  \includegraphics[width=0.5\textwidth]{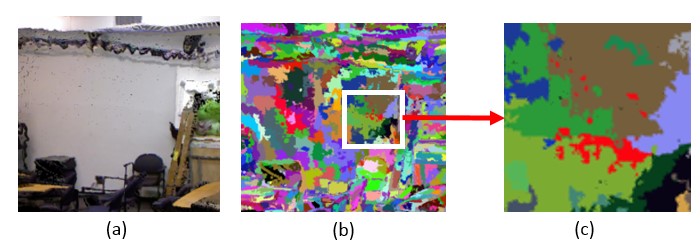}
\caption{Example of dispersed segments. (a) 3D point cloud with RGB color overlay, (b) 3D point cloud with resulting super-points colored in different colors, and (c) enlargement of a region with dispersed segments.}
\label{fig:DispersedSegsExample}
\end{figure}

All the above options excluding the last may lead to undesirable over-segmentation results, including dispersed segments that appear as small islands. These are clusters that are disconnected in the 3D space. Figure \ref{fig:DisperdedSegIllustration} illustrates the cause for this problem. The blue and yellow nodes represent points in two different resulting super-points. The black lines illustrate graph edges of an 8-connected image-based graph, and the red lines represent additional edges used by other graph construction methods. The illustrated segmentation is clearly impossible when using an 8-connected graph, while it is possible for the other graph types.  Indeed, we observed such problematic results during our evaluation (detailed in Section \ref{Sec:Evalutation}). See Figure \ref{fig:DispersedSegsExample} for an example result using a ``Radius R'' based graph. Similar phenomena were encountered for K-NN based graphs and Delaunay based graphs. This issue can be addressed by integrating a connected component analysis stage into the algorithm. However, this limits the control over the number of segments and requires a corresponding image. This problem can be reduced by choosing lower values of K and R.
The advantages and drawbacks of each of the graph construction methods presented above are summarized in Table \ref{table:GraphProsNCons}.

\begin{table*}[tb]
	\centering	
	\tabcolsep = 0.01\textwidth 

	\begin{tabular}{| m{0.2\textwidth} | m{0.33\textwidth} | m{0.33\textwidth}|}
	\hline

	 \centering\textbf{Graph Construction Method} &
     \centering\textbf{Advantages} &
     \centering\textbf{Drawbacks}
     \tabularnewline
    \hline
	Radius R\ &
            \begin{itemize}[label=\textbullet]
              \item Integrated edge length limitation
             \end{itemize}
             \ &
            \begin{itemize}[label=\textbullet]
                 \item R parameter adjustment
                 \item Sensitive to non-uniform point density
                 \item Dispersed segments
            \end{itemize}
            \\
	\hline
    K nearest neighbors\ &
                 \begin{itemize}[label=\textbullet]
                    \item Balanced graph (non-biased to number of neighboring points).
                    \item Insensitive to non-uniform point density
                 \end{itemize}
             \ &
                 \begin{itemize}[label=\textbullet]
                     \item K parameter adjustment
                     \item Distant point connections
                     \item Dispersed segments
                 \end{itemize}
            \\
    \hline
    Delaunay\ &
                  \begin{itemize}[label=\textbullet]
                    \item Insensitive to isolated points
                    \item Parameter independent
                 \end{itemize}
             \ &
                 \begin{itemize}[label=\textbullet]
                     \item Sensitive to noise
                     \item Distant point connections
                     \item Dispersed segments
                 \end{itemize}
            \\
    \hline
    2D Image grid based\ &
                 \begin{itemize}[label=\textbullet]
                    \item Fast
                 \end{itemize}
             \ &
                 \begin{itemize}[label=\textbullet]
                     \item Requires a mapping from a point cloud to a 2D grid.
                     \item Does not directly incorporate 3D information.
                 \end{itemize}
            \\
    \hline
	\end{tabular}

	\caption{Pros and cons of the graph construction methods}
	\label{table:GraphProsNCons}
\end{table*}
Given the advantages and disadvantages above, it remains difficult to conclude which graph construction method is preferable.  Therefore, a quantitative comparison is necessary. This comparison is presented in Section \ref{SubSubSec:GraphConAnsys}.
    \subsection{Descriptor Estimation}
    \label{SubSec:Descriptors}
    The LV algorithm, having been designed for 2D images, uses only RGB color to characterize each pixel.
When the input is a 3D point cloud, we would like to exploit the additional available dimension and to describe also the local geometry of each point. However, it is not easy to characterize the local geometry of a point in a cloud and most studies that attempted to do so yielded ambiguous results. Many of these studies were subsequently surveyed and evaluated in  \cite{Guo2015}, who concluded that some descriptors were better than others for certain tasks while the FPFH \citep{Rusu2009d} descriptor performed well for all of the evaluated tasks. We test here the effectiveness of the FPFH descriptor for the task of over-segmentation and compare it to using the estimated normal at each point. We also test the effectiveness of using the absolute location of each point. Along with color, that gives us 4 different point descriptors, and different combinations thereof. Each graph vertex (3D point) $v_i \in V$  is therefore described by one or more of the following properties:
\begin{enumerate}
\item RGB color $(R_i,G_i,B_i)$
\item Location $(X_i,Y_i,Z_i)$
\item FPFH descriptor vector $(FPFH_i,1...33)$
\item Estimated normal vector $((N_x)_i,(N_y)_i,(N_z)_i)$
\end{enumerate}
Note that most 3D geometrical descriptors are estimated using their local neighbors (even the ``simple'' normal vector). As such, some``impurity'' is introduced into the algorithm: the similarity now relies on all of the neighboring points rather than only on one.

Next we describe how the graph edge weights are derived from these properties. A quantitative comparison between properties and weight alternatives is presented in Section ‎\ref{SubSubSec:WeightsAnsys}.

    \subsection{Edge Weight Assignment}
    \label{SubSec:EdgeWeight}
    As discussed in Section \ref{SubSec:Descriptors}, four different ``modalities'' were considered for describing each point. Respectively, we define the dissimilarity between every two points connected by an edge $e_{ij}$ in the connectivity graph $G$ by:
    \begin{enumerate}
        \item The difference in color as the Euclidean distance in the RGB space,

        \[{w_c}({e_{ij}}) = \frac{{\sqrt {{{({R_i} - {R_j})}^2} + {{({G_i} - {G_j})}^2} + {{({B_i} - {B_j})}^2}} }}{{\sqrt 3 }}.\]

        \item The normalized Euclidian distance in the 3D space,

        \[{w_d}({e_{ij}}) = \frac{{\sqrt {{{({X_i} - {X_j})}^2} + {{({Y_i} - {Y_j})}^2} + {{({Z_i} - {Z_j})}^2}}  - {d_{\min }}}}{{{d_{\max }} - {d_{\min }}}}\]

        where  $d_{\min},d_{\max}$ are the minimum and maximum distances within the given graph.

        \item A measure of planarity using the angle between the estimated normal vectors,

         \[{w_n}({e_{ij}}) = 1 - {\hat N_i} \cdot {\hat N_j}.\]

        \item The histogram intersection \citep{Barla2003} of the FPFH descriptors \citep{Rusu2009d},

        \[{w_{_{FPFH}}}({e_{ij}}) = 1 - \sum\limits_{l = 1}^{33} {\min (FPF{H_{i,l}},FPF{H_{j,l}})}.\]
    \end{enumerate}

As we explain in the next section, the aforementioned dissimilarity measure can be combined into one scalar, by for example, linear combination, and used as the weight on the graph's edge.  Then, one criterion can be used to decide whether to merge two super-points. Another option is to assign a vector of weights to each edge, associated with all, or some, of the dissimilarity measures. In this case, the merge decision will use several criteria.

The decision as to whether the weights should be combined or used separately is left for the next stage. Therefore, the output of the current algorithm stage is a vector of weights ${\bar w_{ij}}$. As explained in Section \ref{SubSec:EXeval}, we tested the contribution of each modality by combing the weights in different ways, as follows:
\begin{enumerate}
  \item Color differences only, as in the original LV,
  ${\bar w_{ij}} = {w_c}({e_{ij}})$
  \item Color differences and Euclidian distance,
  ${\bar w_{ij}} = \left( {{w_c}({e_{ij}}),{w_d}({e_{ij}})} \right)$
  \item Color diffrences and normal differences,
  ${\bar w_{ij}} = \left( {{w_c}({e_{ij}}),{w_n}({e_{ij}})} \right)$
  \item Euclidean distance and normal differences,
  ${\bar w_{ij}} = \left( {{w_d}({e_{ij}}),{w_n}({e_{ij}})} \right)$
  \item Color differences and FPFH descriptor differences,
  ${\bar w_{ij}} = \left( {{w_c}({e_{ij}}),{w_{_{FPFH}}}({e_{ij}})} \right)$
  \item Color differences, Euclidean distance and normal differences,
  ${\bar w_{ij}} = \left( {{w_c}({e_{ij}}),{w_d}({e_{ij}}),{w_n}({e_{ij}})} \right)$
\end{enumerate}

The following section addresses the question of how to use these weight vectors when sequentially deciding whether two super-points should be merged.

    \subsection{Merge Criterion}
    \label{SubSec:MergeC}
    In the original LV algorithm \citep{Felzenszwalb2004, Felzenszwalb1998}, subgraph merging begins by sorting the connectivity graph edges by the weight defined only by the color differences. Then, the connectivity graph edges are traversed in ascending order and a merge decision is made. Let $e_{ij}$ be the graph edge that connects the two sub-graphs $C_{i}$  and $C_{j}$. Let  $w(e)$ be the (scalar) weight on edge $e$, and let  $MST(C_x)$ be the minimum spanning tree of $C_x$. The merge criterion checks whether

    \[w({e_{ij}}) \le \mathop {\min }\limits_{x\, \in \,\{ i,j\} } \,\,\,\,\left( {\mathop {\max }\limits_{e\, \in \,MST({C_x})} w(e)\,\, + \frac{\delta }{{\left| {{C_x}} \right|}}} \right).\]

    Note that $\frac{\delta }{{\left| {{C_x}} \right|}}$ is a segment-dependent adaptive threshold function in which $\delta $  is a user controlled parameter that is related to the desired number of output segments.

    In our work, a few different weights associated with the different modalities are considered for the merge criterion. The merge criterion is the decision whether or not to merge two sub-graphs based on a comparison between the weight on the connecting graph edge and an adaptive threshold value. Integrating the weights is a non-trivial task, commonly described as the problem of ``adding apples to oranges''. Combining them into one scalar requires each of the weights to be normalized and factorized using parameters that require adjustment. Furthermore, some information may be lost in this process. For example, two distant points on different objects may have similar color and collinear normal vectors; when combining these weights the Euclidean distance's influence is averaged with the other weights' influence and the effective weight may falsely imply that both points belong to the same segment. On the other hand, performing separate comparisons for each modality violates some of the algorithm's original assumptions and guarantees. Normalization is also required  in this case in order to reduce the number of adjustable parameters.

    In this work we compare the following options for using the weight information:
    \begin{enumerate}
      \item Linearly combining all modalities to a scalar weight and using a single merge criterion,
      \[w({e_{ij}}) = {k_c}{w_c}({e_{ij}}) + {k_d}{w_d}({e_{ij}}) + {k_n}{w_n}({e_{ij}}),\]
       \[w({e_{ij}}) \le \mathop {\min }\limits_{x\, \in \,\{ i,j\} } \,\,\,\,\left( {\mathop {\max }\limits_{e\, \in \,MST({C_x})} w(e)\,\, + \frac{\delta }{{\left| {{C_x}} \right|}}} \right).\]
      \item Checking separate criteria for each modality and merging only if all criteria are met. If, for instance, we test the use of the weights associated with color, normal, and Euclidean distance, then the criteria that will be checked are:
          \[\begin{array}{l}
\,\,\,\,\,\,\,\,\,\,\,\,{w_c}({e_{ij}}) \le \mathop {\min }\limits_{x\, \in \,\{ i,j\} } \,\,\,\,\left( {\mathop {\max }\limits_{e\, \in \,MST({C_x})} {w_c}(e)\,\, + \frac{\delta }{{\left| {{C_x}} \right|}}} \right)\\
and\,\,\,\,\\
\,\,\,\,\,\,\,\,\,\,\,\,{w_d}({e_{ij}}) \le \mathop {\min }\limits_{x\, \in \,\{ i,j\} } \,\,\,\,\left( {\mathop {\max }\limits_{e\, \in \,MST({C_x})} {w_d}(e)\,\, + \frac{\delta }{{\left| {{C_x}} \right|}}} \right)\\
and\,\,\,\,\\
\,\,\,\,\,\,\,\,\,\,\,\,{w_n}({e_{ij}}) \le \mathop {\min }\limits_{x\, \in \,\{ i,j\} } \,\,\,\,\left( {\mathop {\max }\limits_{e\, \in \,MST({C_x})} {w_n}(e)\,\, + \frac{\delta }{{\left| {{C_x}} \right|}}} \right)
\end{array}.\]
    \end{enumerate}

As mentioned above, the number of segments is controlled indirectly by adjusting the δ parameter that intuitively reflects the initial tolerance of point dissimilarity within a segment. It is possible to use a different parameter for each of the modalities; we decided not to do so in order not to introduce another parameter adjustment and tuning challenge that would make it much more difficult to control the number of output segments.

    \subsection{Over-segmentation Post-processing}
    \label{SubSec:OS_PostP}
In LV, as well as in all the algorithms tested here, a post-processing stage merges all the small segments with the segment closest to them in the graph. Small segments are defined as segments with less than $10\%$ of the number of points there would have been if the 3D point cloud had been subdivided uniformly (number of points divided by the desired number of segments). Note that for some applications the current post-processing stage can be replaced by omitting small segments.

\section{Evaluation of LV 3D Variants }
\label{Sec:Evalutation}
This section details the experimental results of the family of point cloud over-segmentation algorithms that extended the 2D LV algorithm to 3D. We first provide a summary of the common evaluation metrics for over-segmentation. Then, each algorithm stage is thoroughly analyzed, using the original LV as the baseline. We then select the best choice for each stage. Afterwards, we use this analysis to derive the PCLV algorithm. We compare its performance to that of other state-of-the-art 2D and 3D over-segmentation algorithms in Section \ref{Sec:PerformanceAnalysis}.

The performance analysis was performed on  the NYU Depth V2 \citep{Silberman:ECCV12} dataset, which contains 1449 images of indoor scenes acquired by Microsoft Kinect. It includes RGB images aligned with depth intensity images, human labeled ground truth segmentation, camera parameters for 3D point cloud reconstruction, and additional data. For this dataset the normal vectors provided by \cite{Ladicky2014} were used.

    \subsection{Evaluation Metrics}
    \label{SubSec:EvalMetrics}
    The dominant metrics used to test and compare the performance of over-segmentation algorithms are \textit{boundary recall} and \textit{under-segmentation error} \citep{Neubert2012}.  We use the boundary recall  definition from \cite{Martin2001a} and the under-segmentation error definition from \cite{Silberman:ECCV12}. Boundary recall quantifies the fraction of the ground truth boundaries that intersect with the algorithm's output boundaries. The under-segmentation error measures the area of incorrect segment overlaps. In 2D over-segmentation studies these metrics were defined and used mainly for evaluating and comparing algorithms using the Berkeley Segmentation Dataset and Benchmark \citep{Martin2001a}. Generally, such an evaluation is only possible when an annotation of the borders between objects is available. Border annotation directly on a 3D point cloud is a difficult task. For the KITTI point-clouds dataset \citep{Geiger2013}, annotations are available for some of the objects (people and vehicles), but full segmentation of scenes is not yet provided. We follow \cite{Papon2013} and project the 3D points to a 2D domain; we then compare the algorithm's results to the manually annotated segmentation provided by the NYU dataset for that projection.

    The evaluation metrics mentioned above will always prefer many small segments, and will provide the best scores for the trivial over-segmentation, where each point is considered as a separate segment. This is obviously an undesirable result. The objective is, of course, to get the highest boundary recall and the lowest over-segmentation error with a minimal number of segments. Most over-segmentation algorithms allow some control over the number of output segments. Therefore, each metric is evaluated for each algorithm as a function of the number of output segments.
    The formulation of the measures is detailed below.

        \subsubsection{Boundary Recall}
        \label{SubSubSec:BRecall}
        This metric evaluates the number of boundary pixels correctly labeled by the algorithm versus the total number of boundary pixels. It is defined to be the fraction of ground truth ($GT$) edges that fall within a certain distance $d$ of at least one super-pixel boundary. Given a ground truth boundary image $T$ and the algorithm output boundary image $B$, the number of true positive ($TP$) pixels is computed by summing all boundary pixels in $T$ for which exists a boundary pixel in $B$ within a range $d$. The number of false negative ($FN$) pixels is computed by summing the boundary pixels in $T$ for which does not exist a boundary pixel in $B$ within a range $d$. We used $d = 2$. Finally, the boundary recall (BR) is computed:
        \[BR = \frac{{TP}}{{TP + FN}}.\]

        \subsubsection{Under-segmentation Error}
        \label{SubSubSec:UE}
        The under-segmentation error ($UE$) evaluation metric penalizes the algorithm on segments that cross ground-truth borders, by considering the ``trespassed'' areas as errors. Let $S$ be a ground-truth segment, let $P$ be an intersecting output segment, let $P_{in}$  denote the intersection $P \cap S$, and let $P_{out}$  denote  $P \cap \bar S$   (the points in P that do not intersect with $S$). Either  $P_{in}$ or $P_{out}$ has the smallest number of points, and, for each $S$ and $P$, the metric will integrate that number into the total error.

        \[UE = \frac{1}{N}\left[ {\;\mathop \sum \limits_{S \in GT} \left( {\mathop \sum \limits_{P:P \cap S \ne \emptyset } {\rm{min}}\left( {|{P_{in}}\left| {,|{P_{out}}} \right|} \right)} \right)\;} \right]\]
        where $N$  is the number of segments in the $GT$ image.
        Note that this formulation distinguishes between internal and external clusters. An internal cluster is a cluster with a majority of points within $S$. This formulation prevents over-penalizing small ``bleeding'' overlaps.

    \subsection{Experimental Evaluation}
    \label{SubSec:EXeval}
    In this section the different extensions are analyzed with respects to the evaluation metrics.
        \subsubsection{Graph Construction Analysis}
        \label{SubSubSec:GraphConAnsys}
\begin{figure}[tb]
	\begin{center}
  \subfloat[ ]{\includegraphics[width=0.45\textwidth]{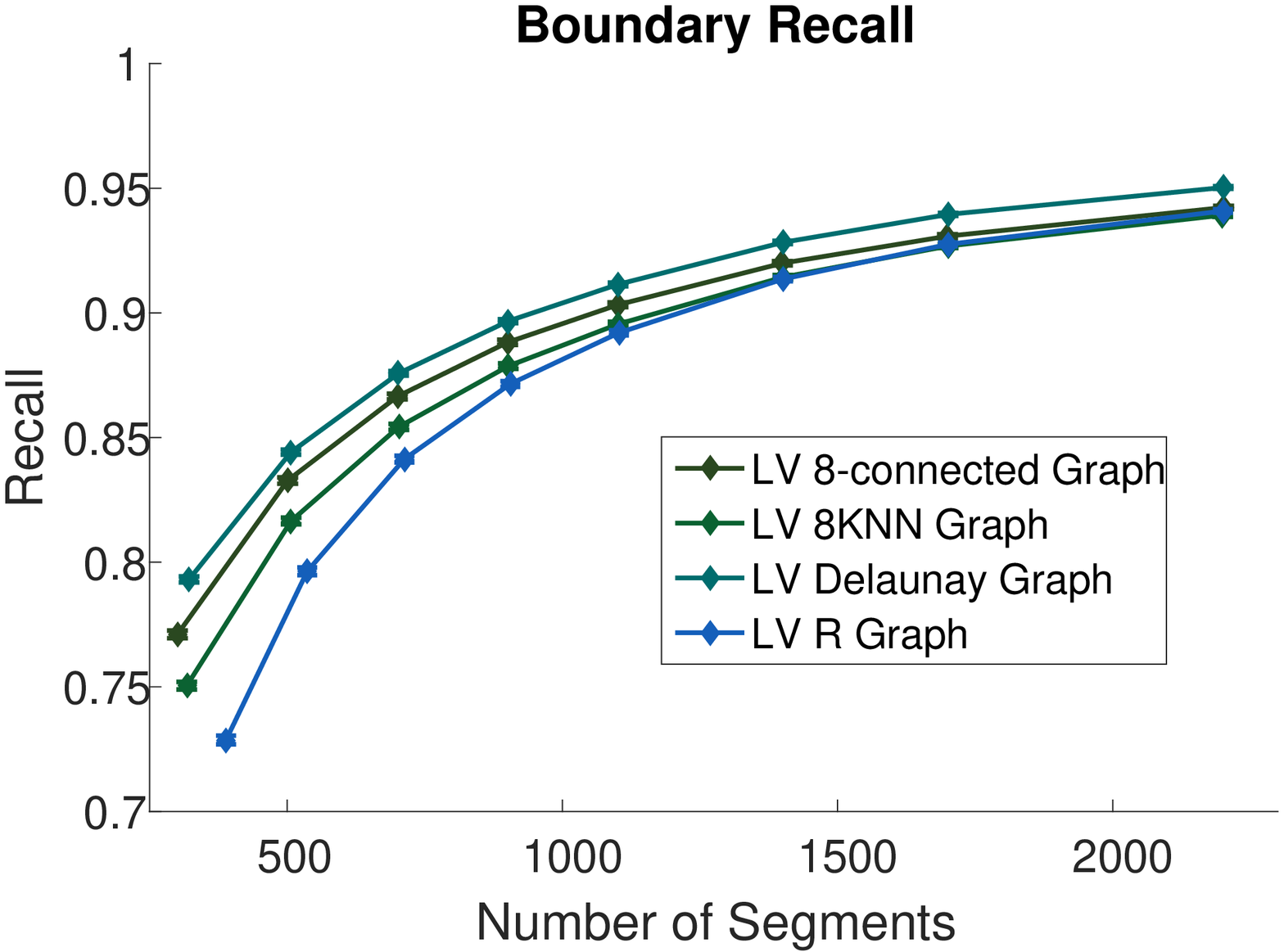}}
  \newline
  \subfloat[ ]{\includegraphics[width=0.45\textwidth]{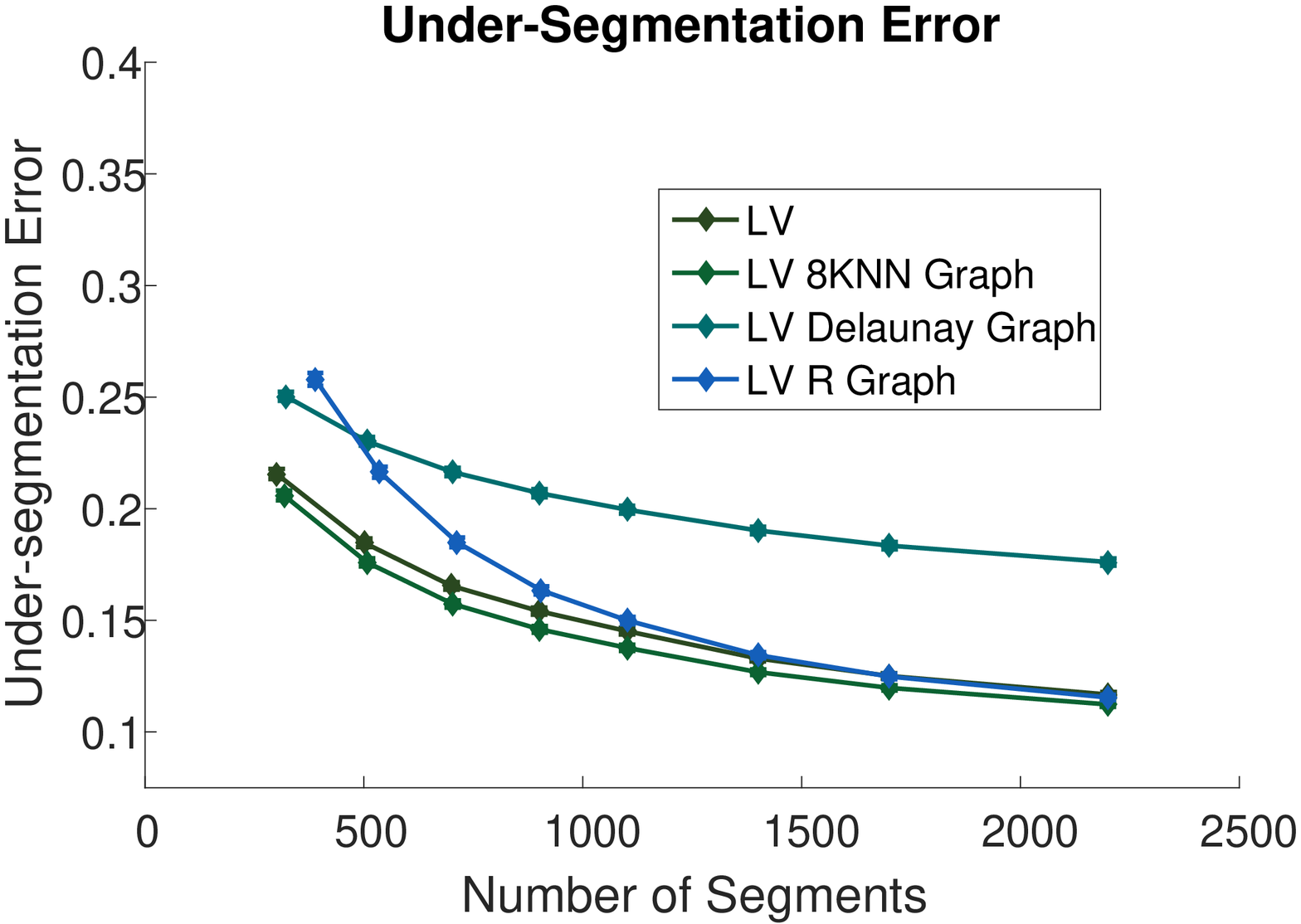}}
	\end{center}
	\caption{ Graph construction comparison. (a) Boundary recall; (b) Under-segmentation error.}
	\label{fig:GraphConCompare}
\end{figure}

        The graph construction method may vary by application. Here we compared the results obtained using the four graph construction methods presented in Section ‎\ref{SubSec:GraphCon}  while maintaining all of the original LV stages. This was done in order to isolate the effect of the graph construction choice. The results can be seen in Figure         \ref{fig:GraphConCompare}.

Surprisingly, integrating the third-dimension information into the graph construction stage does not always improve the algorithm's performance, as can be seen in the results for the KNN graph and the radius based graph. Furthermore, for the Delaunay graph, while the recall has improved, this improvement comes with a significant increase in the under-segmentation error. Taking into account these results and the advantages and disadvantages of each method, as summarized in Table \ref{table:GraphProsNCons}, we conclude that the best choice for this stage would be to use the 8-connected graph construction method when a corresponding image is available and to use a KNN graph when it is not.

        \subsubsection{Descriptors and Edge Weight Assignment }
        \label{SubSubSec:WeightsAnsys}

\begin{figure}[tb]
            	\begin{center}
                  \subfloat[ ]{\includegraphics[width=0.45\textwidth]{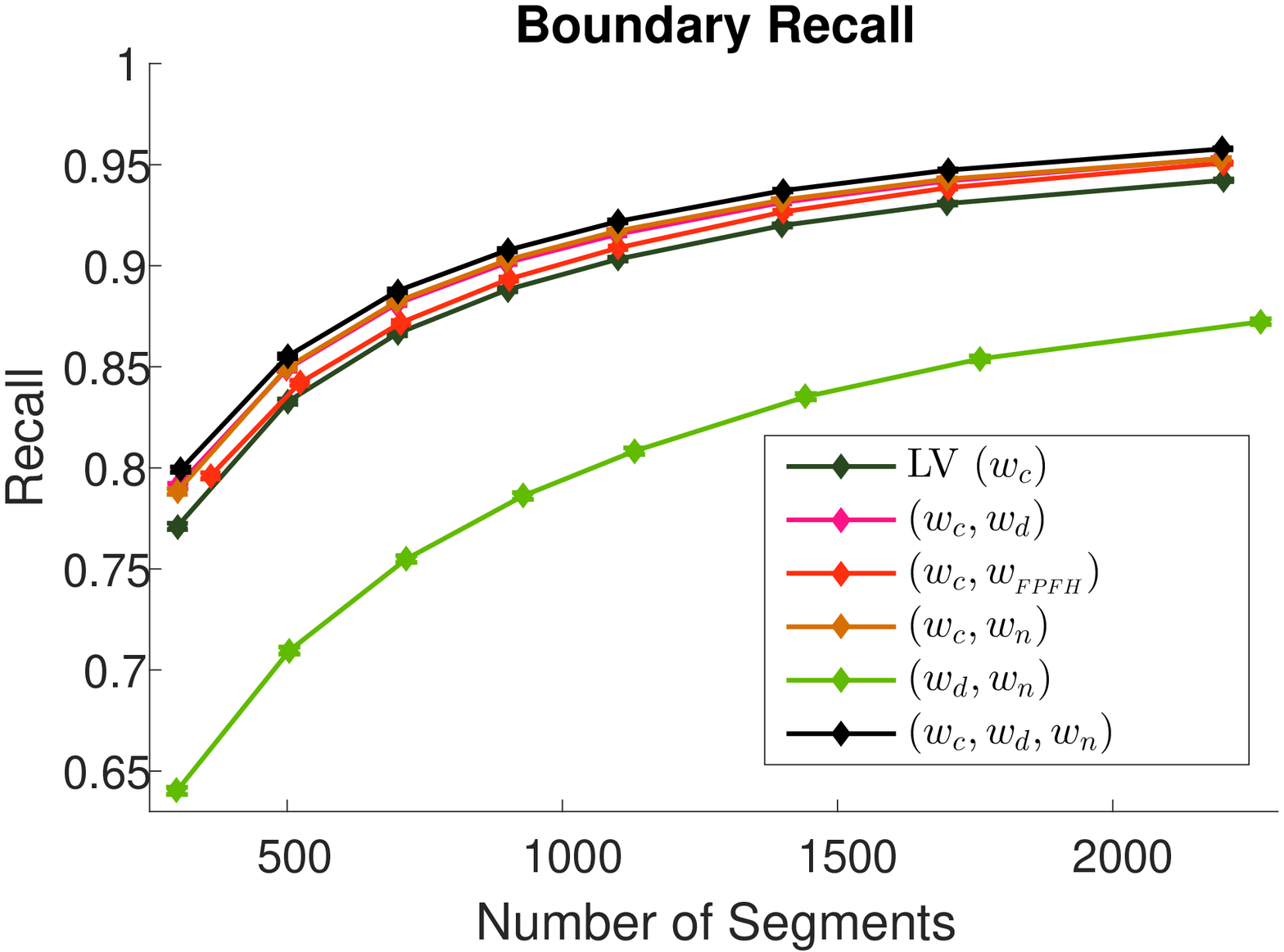}}
                  \newline
                  \subfloat[ ]{\includegraphics[width=0.45\textwidth]{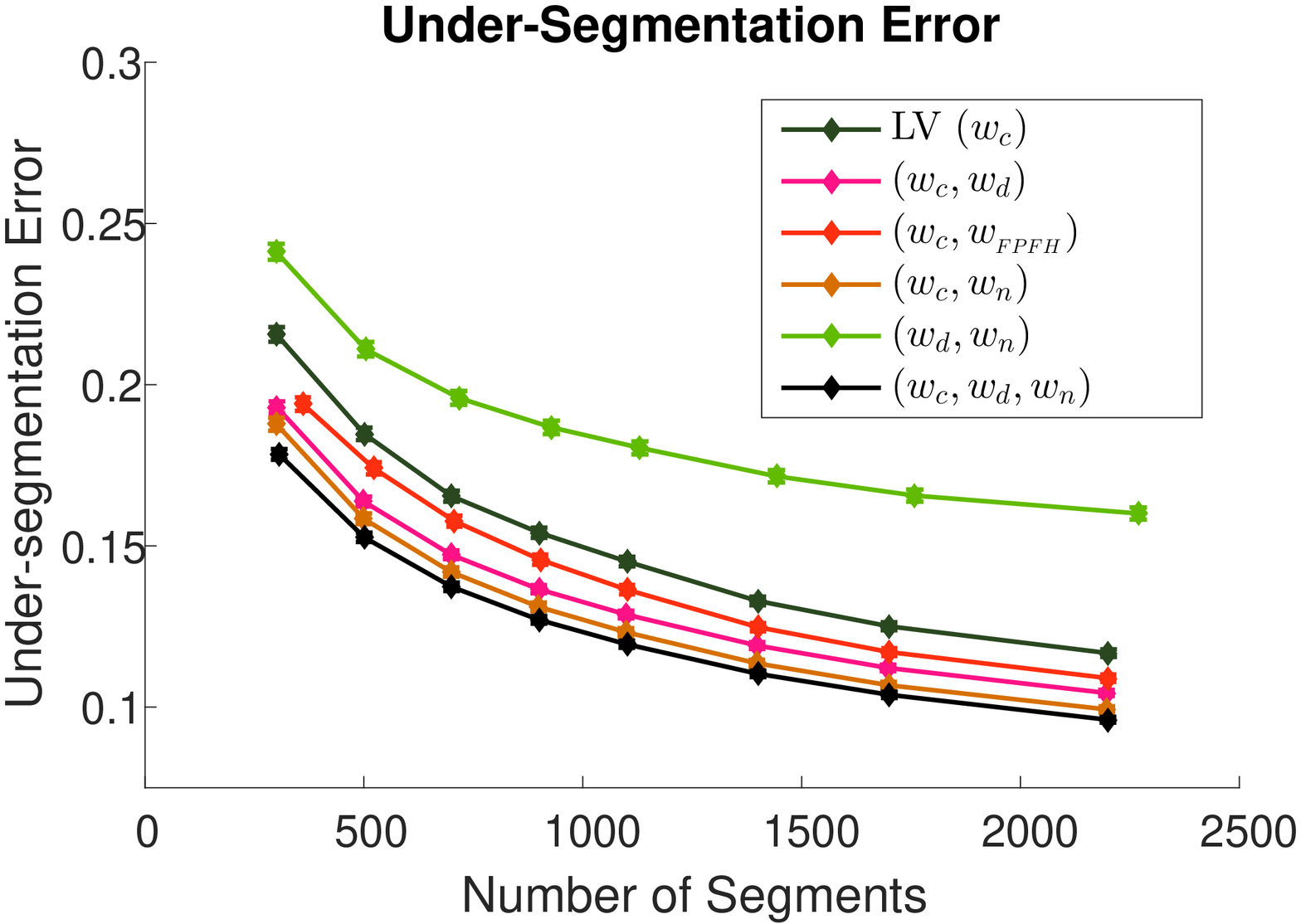}}
            	\end{center}
        	\caption{ Boundary recall (a) and under-segmentation error (b) for the 4 modalities: color ($w_c$), distance ($w_d$), normal vectors ($w_n$), FPFH feature ($w_{_{FPFH}}$), and their combinations.}
        	\label{fig:ModalityCompare}
        \end{figure}

        The modalities available for 3D point clouds may vary according to the application. Here we compared the results of algorithms using each of the four modalities defined in Section \ref{SubSec:EdgeWeight}. We isolated the effect of descriptor and weight choice by setting the graph construction to be the same as in the LV algorithm (8-connected) and applying the merge criterion separately on each modality using a single adjustable parameter for the adaptive threshold. Figure \ref{fig:ModalityCompare} depicts the boundary recall and under-segmentation error. It can be seen that the recall is substantially lower and the error substantially higher for the variant that does not use the color information. Therefore, contrary to the conclusion of \cite{Karpathy2013}, color clearly plays an important role and each modality further contributes to performance. Surprisingly, FPFH contributed less to performance than did the angle between normal vectors. This is because the FPFH estimation relies heavily on the surrounding points; therefore, adjacent points will have similar FPFH values.  Furthermore, it can be seen that combining the color, normal, and Euclidian distance yielded the best over-segmentation performance.

\begin{figure}[tb]
               \begin{center}
               \captionsetup[subfigure]{labelformat=empty}
                  \subfloat{\includegraphics[width=0.23\textwidth]{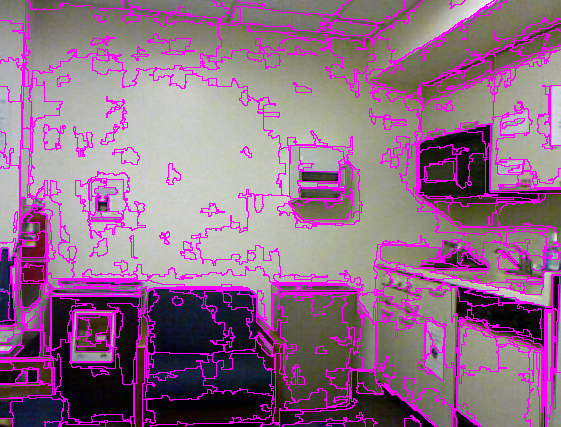}}
                  \hspace{0.01\textwidth}
                  \subfloat{\includegraphics[width=0.23\textwidth]{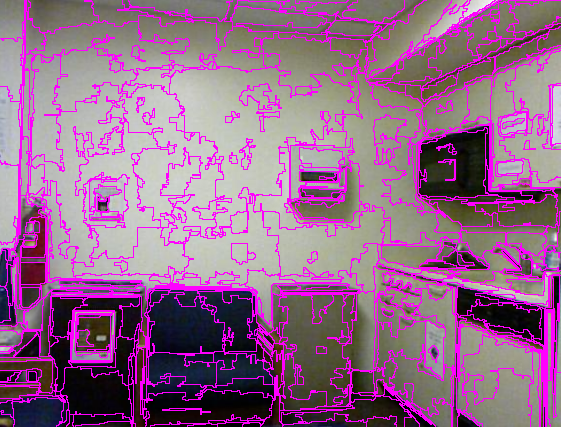}}
                  \newline
                  \subfloat[LV ($w_c$)]{\includegraphics[width=0.23\textwidth]{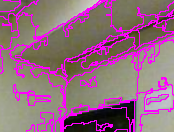}}
                  \hspace{0.01\textwidth}
                  \subfloat[($w_c, w_d, w_n$)]{\includegraphics[width=0.23\textwidth]{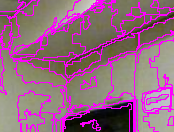}}
            	\end{center}
            \caption{ The advantage of using normal vectors. Top: over-segmentation results; bottom: enlarged region of interest; left: using color ($w_c$); right: using a combination of color, distance, and normal vectors ($w_c, w_d, w_n$).}
        	\label{fig:OSNormalAdvImg}
        \end{figure}

        An additional qualitative comparison between the over-segmentation results is presented in Figure \ref{fig:OSNormalAdvImg} (where the number of output segments is 500). At the top right corner of the image, a shaded intersection between the wall and the ceiling can be seen; a zoomed-in view of the over-segmentation results on this region is also shown. The shading makes it impossible to find the location of this intersection using color information alone because the color is approximately uniform. This is evident in the results of the LV algorithm, where the border follows the shade line rather than the underlying intersecting corner, while for the extension that uses 3D information the corner boundary was correctly located. This example emphasizes the contribution of the normal vector to the over-segmentation process.

        \subsubsection{Merge Criteria Selection and Further Comparison to Previously Suggested LV Extensions }
        \label{SubSubSec:MergeCAnsys}
        As discussed in Section \ref{SubSec:MergeC}, it is possible to linearly combine the modality-based weights into one weight and then use one merge criterion, or use multiple criteria, one for each modality, all of which must be met if a merge decision is to be made. We compared these two options. Following the results reported in Section \ref{SubSubSec:WeightsAnsys}, we used the modalities of color, normal and Euclidean distance for both options. The results are reported in Figure 7. The dark purple lines depict the performance of the linear combination option, while the black lines present the results of the multiple criteria option.  The latter option is more successful, while the linear combination option performs slightly worse than the original LV algorithm. This finding concurs with our notion that combining the weights will average significant differences in color space.
        As discussed in Section \ref{SubSec:3D_OS}, point-cloud extensions for LV were suggested in \cite{Strom2010,Karpathy2013}, and \citep{Schoenberg2010}. We also compared our approach to the approaches suggested in these works. The approach of \cite{Schoenberg2010} is similar to the linear combination option discussed above.The method of \cite{Karpathy2013}, which does  not exploit the color information, has lower recall and higher under-segmentation error than our results. Finally, the method of  \cite{Strom2010}, which yields a single data point on the evaluation graphs, has significantly lower recall and higher under-segmentation error than those of the other approaches. This may be attributed to the use of two predefined parameters for the color and normal vector modalities, which may require optimization and adjustment for every data type.

        \begin{figure}[tb]
            \begin{center}
                  \subfloat[ ]{\includegraphics[width=0.45\textwidth]{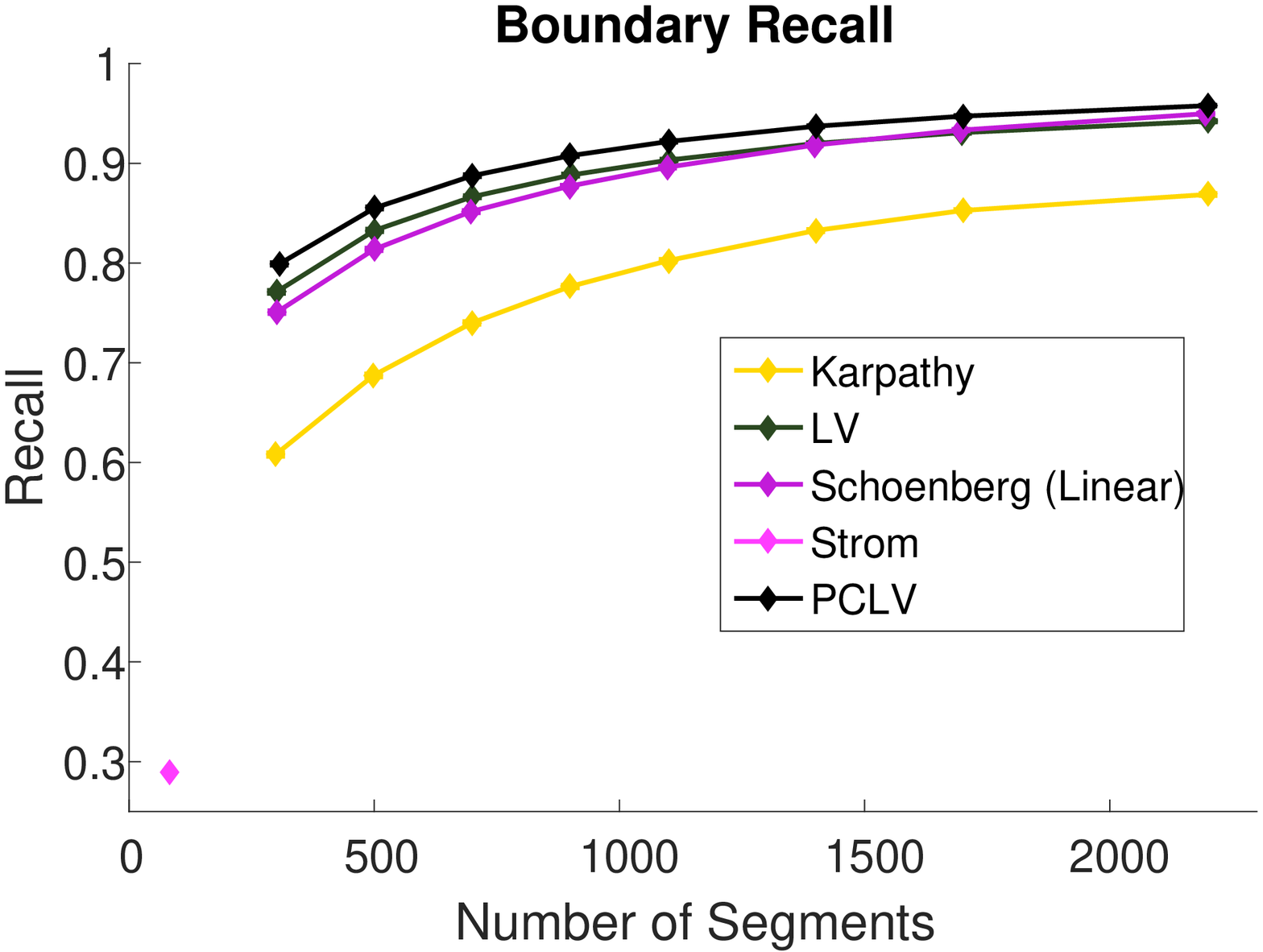}}
                  \newline
                  \subfloat[ ]{\includegraphics[width=0.45\textwidth]{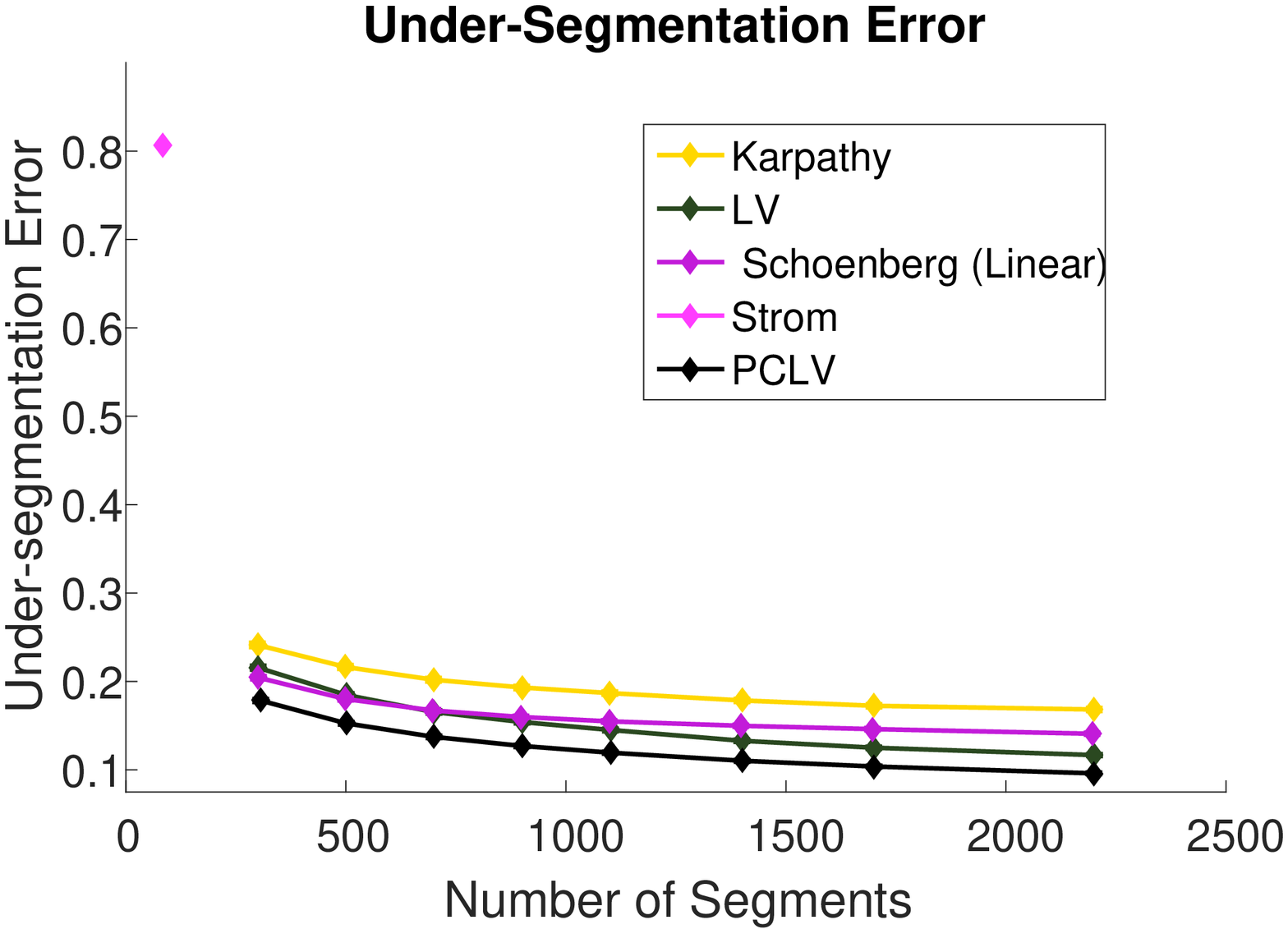}}
            \end{center}
            \caption{ Single vs. multiple merge criteria comparison, and comparison to previously suggested LV extensions. (a) Boundary recall and (b) under-segmentation error.}
        	\label{fig:MergeCCompare}
        \end{figure}

\section{The Proposed Point Cloud Local Variation Algorithm (PCLV)}
\label{Sec:PCLV}
        Based on the analysis in the previous section, it is evident that the choices of an 8-connected graph, a descriptor based on color, normal vectors, and Euclidean distance, and multiple merge criteria yield better performance than all the surveyed extensions. We denote the algorithm that makes those choices PCLV (Point-Cloud Local Variation). The PCLV algorithm is formally summarized in Algorithm \ref{algo:PCLV}.

\begin{algorithm}[tb]
	\caption{Point Cloud Local Variation Algorithm}
	\label{algo:PCLV}
	\textbf{Input:} 3D Point Cloud.\\
	\textbf{Output:} Set of components $C_1,...,C_n$ defining the super-points
	\begin{algorithmic}[1]
        \STATE Construct connectivity graph $G = (V,E)$
        \begin{enumerate}
          \item If an image mapping is available construct an 8-connected graph.
          \item Otherwise construct a K nearest neighbor graph.
        \end{enumerate}
        \STATE Compute descriptors for each point $(X,Y,Z,R,G,B,N_x,N_y,N_z)$
        \STATE Compute and assign multi-value graph weights for each edge $(w_c,w_d,w_n), e \in E$
		\STATE Sort $E$ by non-decreasing edge color weight $\left(e_1, e_2,..., e_m\right)$
		\STATE Initialize segmentation $S^0$ with each vertex being a component
		\FORALL{$q = 1,...,m$}
			\STATE $e_q = (v_i, v_j) \gets$ edge with the $q$th smallest weight
			\STATE $C_i^{q-1} \gets$ component of $S^{q-1}$ containing $v_i$
			\STATE $C_j^{q-1} \gets $ component of $S^{q-1}$ containing $v_j$
            \IF{\[\begin{array}{l}
            \left( \begin{array}{l}
            \left( {\mathop C\nolimits_i^{q - 1}  \ne \mathop C\nolimits_j^{q - 1} } \right)\\
            {\rm{and}}\,\,\,\left( {{w_c}({e_q}) \le \mathop {\min }\limits_{x\, \in \,\{ i,j\} } \,\,\,\,\left( {\mathop {\max }\limits_{e\, \in \,MST({C_x})} {w_c}(e)\,\, + \frac{\delta }{{\left| {{C_x}} \right|}}} \right)} \right)\\
            {\rm{and}}\,\left( {{w_d}({e_q}) \le \mathop {\min }\limits_{x\, \in \,\{ i,j\} } \,\,\,\,\left( {\mathop {\max }\limits_{e\, \in \,MST({C_x})} {w_d}(e)\,\, + \frac{\delta }{{\left| {{C_x}} \right|}}} \right)} \right)\\
            {\rm{and}}\,\left( {{w_n}({e_q}) \le \mathop {\min }\limits_{x\, \in \,\{ i,j\} } \,\,\,\,\left( {\mathop {\max }\limits_{e\, \in \,MST({C_x})} {w_n}(e)\,\, + \frac{\delta }{{\left| {{C_x}} \right|}}} \right)} \right)
            \end{array} \right)\,\\
            \,\,\,\,\,\,\,\,\,\,\,\,
            \end{array}\]
            }
                \STATE $S^q = S^{q-1} \cup \left\{ C_i^{q-1} \cup C_j^{q-1} \right\} \setminus \left\{ C_i^{q-1}, C_j^{q-1} \right\}$
			\ELSE
				\STATE $S^q = S^{q-1}$
			\ENDIF
		\ENDFOR
		\STATE \textbf{Postprocessing:} Merge all small segments to closest neighbor.
	\end{algorithmic}
\end{algorithm}

\section{PCLV Performance Analysis}
\label{Sec:PerformanceAnalysis}
We compared PCLV's performance to that of 2D and 3D state-of-the-art over-segmentation algorithms (that are not LV-based) using the NYU Depth V2 cluttered indoor scenes dataset \citep{Silberman:ECCV12}. In addition, we demonstrate PCLV's performance on outdoor scenes.
    \subsection{Indoor Scene Over-Segmentation Performance Evaluation}
    \label{SubSec:IndoorEval}
\begin{figure}[tb]
            \begin{center}
                  \subfloat[ ]{\includegraphics[width=0.45\textwidth]{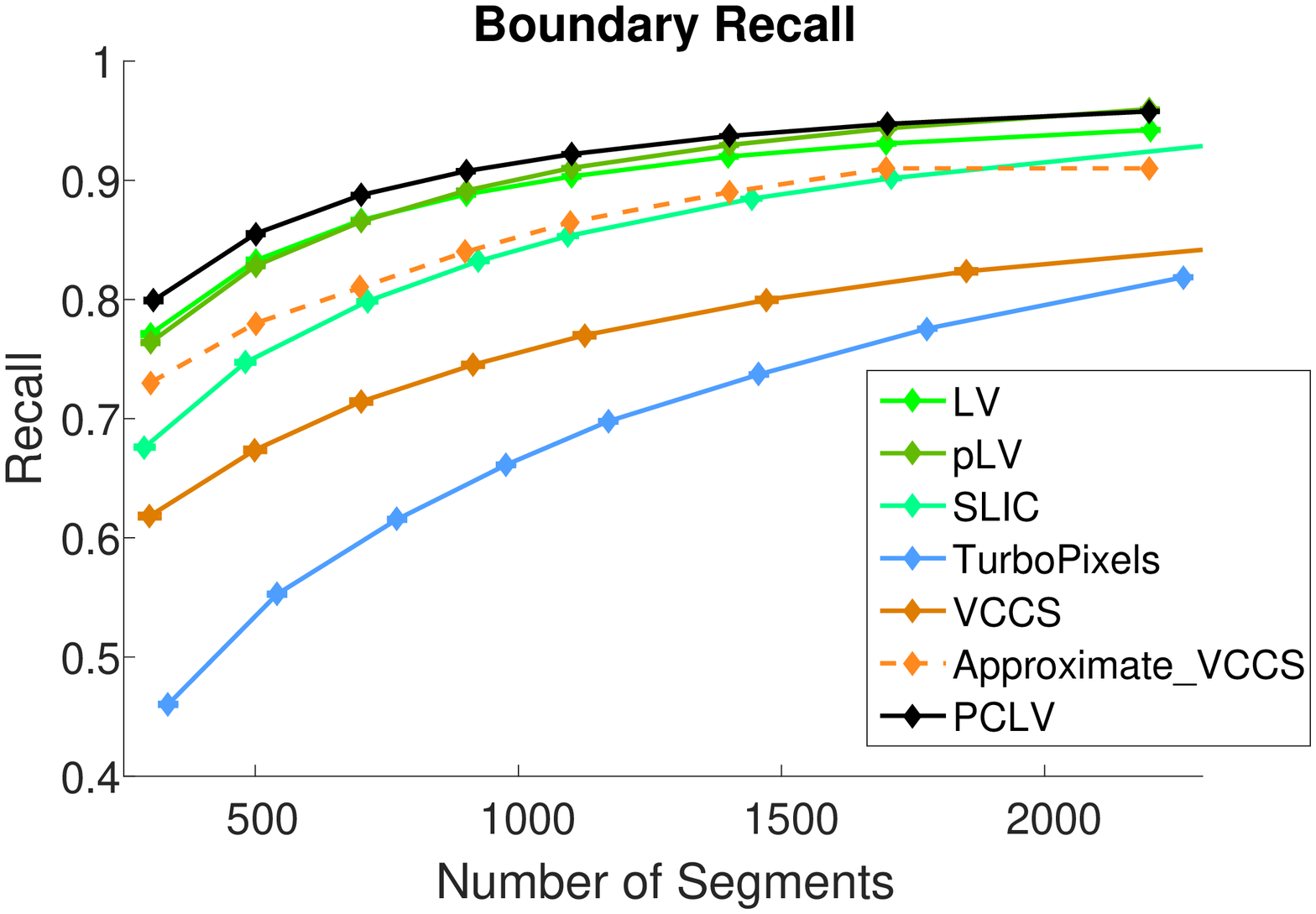}}
                  \newline
                  \subfloat[ ]{\includegraphics[width=0.45\textwidth]{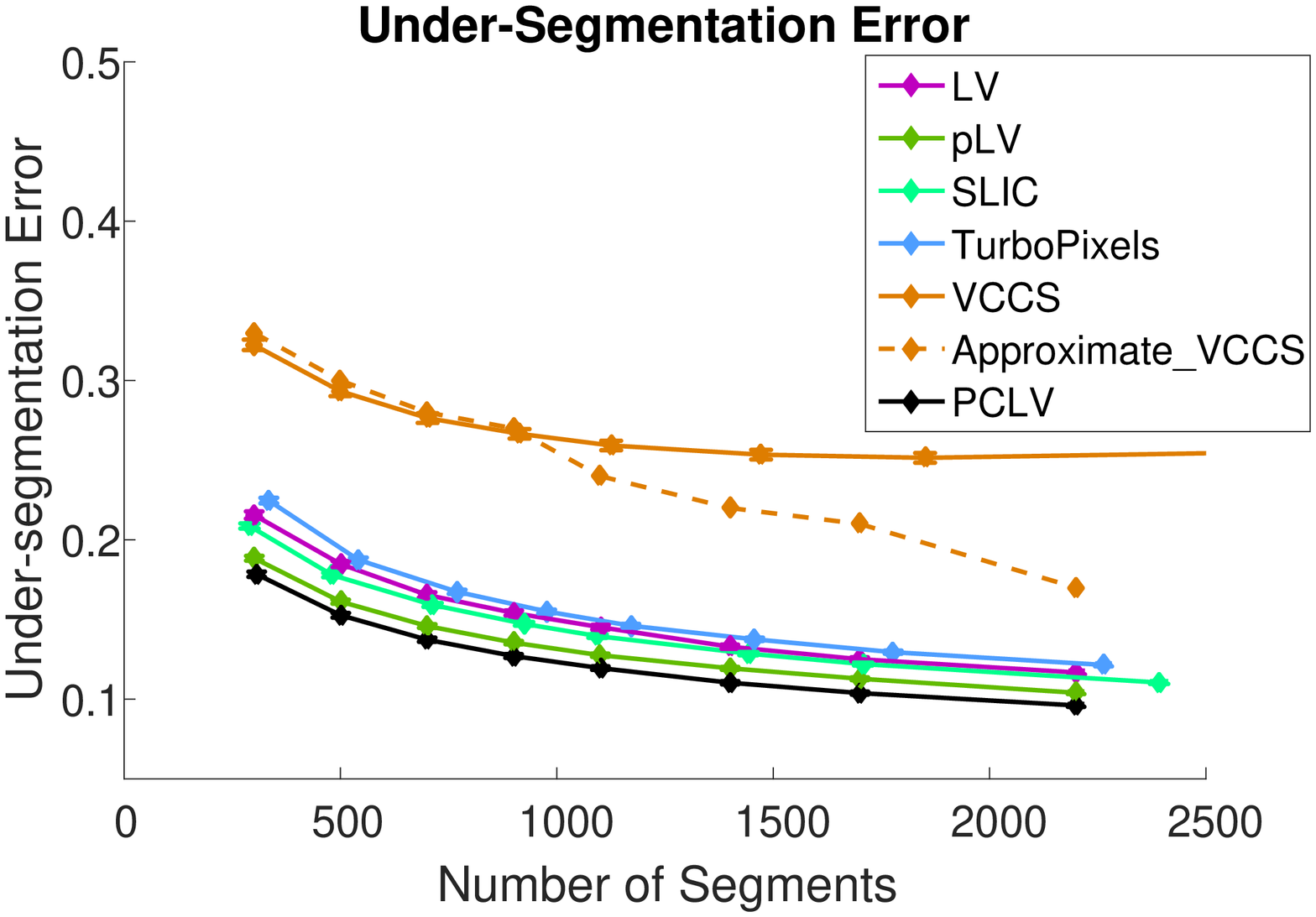}}
            \end{center}
            \caption{ Performance analysis of PCLV compared to 2D and 3D state-of –the-art over-segmentation algorithms. (a) Boundary recall and (b) under-segmentation error as a function of the number of segments.}
        	\label{fig:StateOfTheArtCompare}
        \end{figure}

     Figure \ref{fig:StateOfTheArtCompare} depicts the boundary recall and under-segmentation error results for the following algorithms:  LV \citep{Felzenszwalb1998,Felzenszwalb2004}, SLIC \citep{Achanta2012}, pLV \citep{Baltaxe2015}, TP \citep{Levinshtein2009a}, VCCS \citep{Papon2013} and the proposed PCLV. It can be seen that PCLV, marked in black, outperforms all the other algorithms in both evaluation metrics. The result on the 2D algorithms SLIC, LV, pLV and TP are consistent with previously published results on the Berkeley dataset \citep{Baltaxe2015}. The publicly available implementation of the VCCS algorithm in the Point Cloud Library (PCL) \citep{Rusu2011} yields different results than the ones reported in the original paper \citep{Papon2013}. This difference can be attributed to discrepancies between the paper and the publicly available implementation, such as the discrepancy in the geometrical component: the implementation uses the angle between normal vectors while in the paper the FPFH intersection kernel is used. In addition, the pixels with missing depth information are not segmented in the PCL implementation while in the paper a post-processing step is used to integrate SLIC over-segmentation into these regions. Furthermore, we applied the PCL implementation with the default values for several adjustable parameters. Therefore, for fairness, we present both the PCL implementation results (in green) and the approximated data points for the VCCS algorithm retrieved from the original paper \citep{Papon2013} (dashed orange line).  Note that this comparison may still be somewhat inaccurate with regard to the under-segmentation error since in \cite{Papon2013} a slightly different definition for it was used.

\begin{figure*}
                    \includegraphics[width=\textwidth]{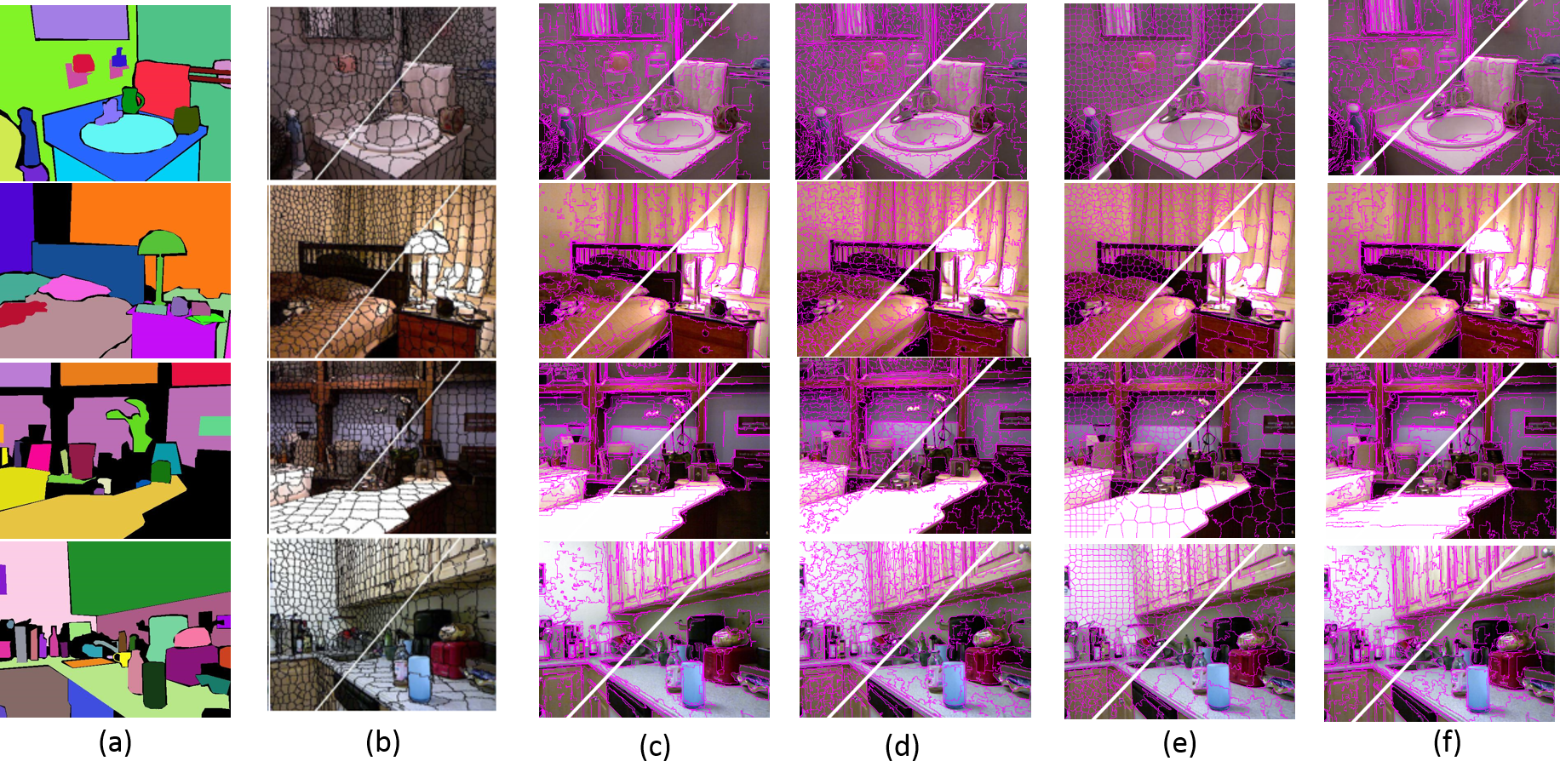}
                    \caption{Results of different over-segmentation algorithms. (a) human labeled ground truth, (b) VCCS, (c) LV, (d) pLV, (e) SLIC, (f) PCLV.}
        	\label{fig:StateOfTheArtCompareVis}
        \end{figure*}

        Figure \ref{fig:StateOfTheArtCompareVis} depicts some examples of the over-segmentation results for the different algorithms when outputting 200 and 1200 segments. The LV, pLV and PCLV algorithms segment the sink as a single unit (top row, (c), (d), and (f)), while the while the  VCCS and SLIC algorithms subdivide it (top row, (b) and (e)). In general, SLIC and VCCS produce regular and uniform segments while LV, pLV and PCLV better follow the GT boundaries. This introduces a compromise between regularity and accuracy \citep{Machairas2015,Veksler2010}. Note that Figure \ref{fig:StateOfTheArtCompareVis} (b) depicts the results for the VCCS algorithm as presented in the original paper.
    \subsection{Outdoor Scene Over-Segmentation Performance Evaluation}
    \label{SubSec:OutdoorEval}
    We also experimented with different outdoor 3D point clouds provided by different industrial companies. We demonstrate here PCLV's performance on a large point cloud acquired by a data capturing vehicle mounted with an interferometry sensor by Geosim, a 3D city modeling technology company \citep{Geosim}.  The scanner is very accurate but the raw point clouds are rather noisy and require some pre-processing.  The pre-processing stage included a basic color threshold filter to remove sensor-generated false points and denoising \citep{Rusu2008c} to remove isolated noisy points. Note that for this data the color information for each point is an intensity value and not RGB. The final input point cloud included approximately 5.75M points. The PCLV algorithm was applied using a $K=8$ nearest neighbor connectivity graph. Normal vectors were estimated using \cite{Hoppe1992}.

\begin{figure}[tb]
            \begin{center}
                \includegraphics[width=0.5\textwidth]{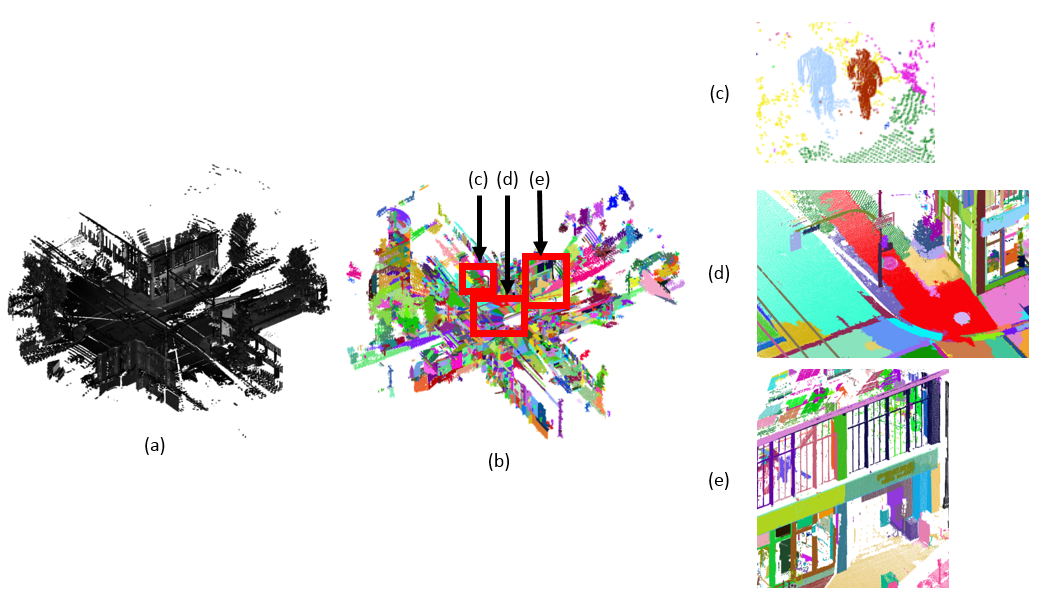}
            \end{center}
            \caption{  Outdoor 3D point cloud and PCLV over-segmentation results. (a) The input point cloud after filtering and denoising, (b) PCLV over-segmentation results for 3500 segments, (c)-(e) enlarged regions of the over-segmented 3D point cloud, where the correctly clustered points are shown for pedestrians (c), stop light poles and sewer covers (d), and building facades and signs (e).}
        	\label{fig:OutdoorResults1}
        \end{figure}

\begin{figure}[tb]
            \begin{center}
                \includegraphics[width=0.5\textwidth]{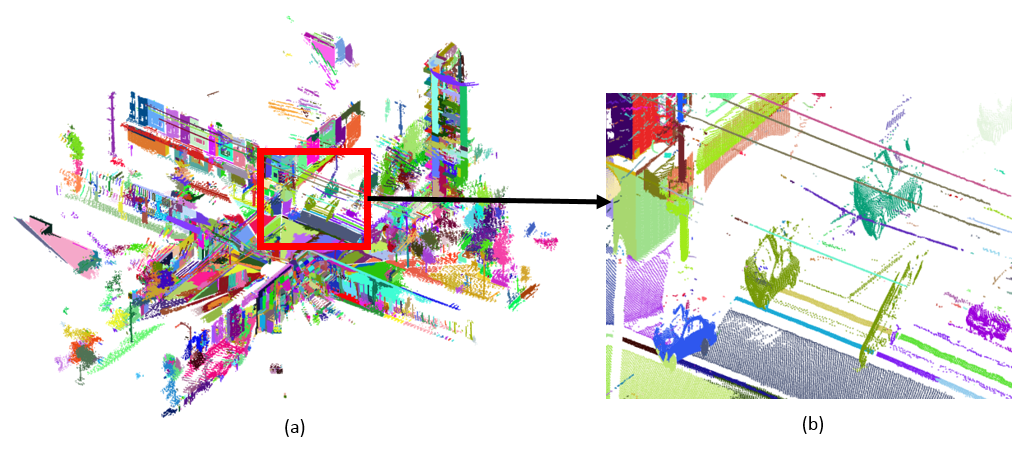}
            \end{center}
            \caption{  PCLV over-segmentation results. (a) The full point cloud from a certain viewpoint, (b) the resulting segments for an enlarged region, showing correctly segmented cars. }
        	\label{fig:OutdoorResults2}
        \end{figure}

     The input 3D point cloud after filtering and denoising is depicted in Figure \ref{fig:OutdoorResults1} (a), and the PCLV algorithm over-segmentation results are depicted in Figure \ref{fig:OutdoorResults1}, (b) which shows each segment in a distinct color. Enlarged regions of the over-segmented 3D point cloud are shown in Figure \ref{fig:OutdoorResults1} (c),(d), and (e), where points located on the same object are clustered together as desired. An additional point of view of this over-segmented cloud is depicted in Figure \ref{fig:OutdoorResults2} (b), where nicely segmented cars can be seen in the enlarged region.


\section{Discussion and Conclusion}
\label{Sec:Discussion}
Super-pixels have become a common preliminary stage in many high level vision related tasks which use 2D images. We introduce a similar notion of super-points. In 3D point clouds, the amount of data is usually much larger than in 2D images, We considered a variety of extensions of the LV algorithm, which is known to be fast and accurate on 2D images. We showed that the extension of LV into the 3D domain is not straightforward for several reasons: (a) Unlike 2D images, 3D point clouds are not structured on a grid, giving rise to the question of how the connectivity graph should be defined; (b) 3D point clouds contain not only color information but also geometric information, giving rise to the question of how to best exploit this additional information; and (c), adapting the LV merge criterion is not trivial when multiple modalities must be taken into account.

After a thorough analysis of these questions, we introduced a new graph-based over-segmentation algorithm for 3D point clouds. Our new Point Cloud Local Variation (PCLV) algorithm is generic in the sense that it is sensor independent and may be applied to any 3D point cloud data.

An extensive empirical comparison of the PCLV algorithm on cluttered indoor scenes from a large Kinect dataset was conducted, along with a qualitative evaluation of the algorithm on outdoor scenes. It was demonstrated that PCLV is highly accurate and outperforms all of the compared 2D and 3D over-segmentation methods. In future work the evaluation may be extended for outdoor data. However, as far as we know, there is currently no benchmark with annotated segmentations for point clouds of outdoor scenes. Therefore future work may also focus on creating such a dataset or extending the annotations for an existing one, such as KITTI \citep{Geiger2013,Niemeyer2014}.

The comparison between over-segmentation methods demonstrated a tradeoff between segment regularity and accuracy. This has been previously addressed in the literature \citep{Veksler2010,Machairas2015}. Regularity has some advantages, such as an approximately equal sized segments, sometimes desirable for a specific application. Its main drawback, however, is the lower correlation with the true segments' boundaries. Furthermore, for the non-regular case, very small segments may be attributed to noise and therefore omitted. This further reduces the size of the output for any following application. Future work may focus on improving the regularity of the PCLV segments.

One of our interesting findings was that more complicated 3D descriptors, such as FPFH, produced less accurate results than those obtained using the simple normal vectors. We believe this is due not to the design of these descriptors but to their being estimated from the data; the more complicated they are, the more bias is introduced into the estimation. Further investigation of this issue will include attempts to integrate different local geometric descriptors in the over-segmentation process, with a focus on the way they are estimated.

\begin{acknowledgements}

This work was supported by Magnet Omek Consortium, Ministry of Industry and Trade, Israel.
\end{acknowledgements}

\end{sloppypar}

%

%


\bibliographystyle{spbasic}      
\bibliography{PCLVPaperRefs}   

%
%

\end{document}